\documentclass[10pt,journal,compsoc]{IEEEtran}
\ifCLASSOPTIONcompsoc
  \usepackage[nocompress]{cite}
\else
  \usepackage{cite}
\fi
\usepackage{subfigure}

\ifCLASSINFOpdf
\else
   \usepackage[dvips]{graphicx}
\fi
\usepackage{url}

\usepackage{graphicx}
\usepackage{color}

\usepackage[linesnumbered,ruled,vlined]{algorithm2e}
\usepackage{amsmath}
\usepackage{threeparttable}
\ifCLASSINFOpdf
\else
\fi

\begin{document}

\title{Adaptive Visibility Graph Neural Network and It’s Application in Modulation Classification}

\author{Qi~Xuan,~\IEEEmembership{Member,~IEEE,}
        Kunfeng~Qiu, Jinchao Zhou, \\
        Zhuangzhi Chen, Dongwei Xu, Shilian Zheng, Xiaoniu Yang
\IEEEcompsocitemizethanks{
\IEEEcompsocthanksitem Q. Xuan, K. Qiu, J. Zhou, Z. Chen, and D. Xu are with the Institute of Cyberspace Security, and also with the College of Information Engineering, Zhejiang University of Technology, Hangzhou
310023, China (E-mail: xuanqi@zjut.edu.cn; yexijoe@163.com;
jinchaozhou73@gmail.com; zzch@zjut.edu.cn; dongweixu@zjut.edu.cn).
\IEEEcompsocthanksitem S. Zheng is with the Science and Technology on Communication Information Security Control Laboratory, Jiaxing 314033, China (E-mail:lianshizheng@126.com).
\IEEEcompsocthanksitem X. Yang is with the Institute of Cyberspace Security, Zhejiang University of Technology, Hangzhou 310023, China, and also with the Science and Technology on Communication Information Security Control Laboratory,
Jiaxing 314033, China (E-mail: yxn2117@126.com).}
}

\markboth{IEEE TRANSACTIONS ON NETWORK SCIENCE AND ENGINEERING}%
{Shell \MakeLowercase{\textit{et al.}}: AVGNet: Adaptive Visibility Graph Neural Network and It’s Application in Modulation Classification}

\IEEEtitleabstractindextext{%
\begin{abstract}
Our digital world is full of time series and graphs which capture the various aspects of many complex systems. Traditionally, there are respective methods in processing these two different types of data, e.g., Recurrent Neural Network (RNN) and Graph Neural Network (GNN), while in recent years, time series could be mapped to graphs by using the techniques such as Visibility Graph (VG), so that researchers can use graph algorithms to mine the knowledge in time series. Such mapping methods establish a bridge between time series and graphs, and have high potential to facilitate the analysis of various real-world time series. However, the VG method and its variants are just based on fixed rules and thus lack of flexibility, largely limiting their application in reality. In this paper, we propose an Adaptive Visibility Graph (AVG) algorithm that can adaptively map time series into graphs, based on which we further establish an end-to-end classification framework AVGNet, by utilizing GNN model DiffPool as the classifier. We then adopt AVGNet for radio signal modulation classification which is an important task in the field of wireless communication. The simulations validate that AVGNet outperforms a series of advanced deep learning methods, achieving the state-of-the-art performance in this task.
\end{abstract}

}

\maketitle

\IEEEdisplaynontitleabstractindextext

\IEEEpeerreviewmaketitle

\ifCLASSOPTIONcompsoc
\IEEEraisesectionheading{\section{Introduction}\label{introduction}}
\else
\section{Introduction}
\label{introduction}
\fi

\IEEEPARstart{G}{raphs} and time series are two typical kinds of data in the real world. The former is used to describe the topological structure of many complex systems, where nodes represent subsystems and links capture the relationship between them, leading to the emergence of complex networks~\cite{watts1998collective,barabasi1999emergence}; while the latter is used to capture the temporal dynamics of these systems. At present, biology, social science, computer science and other disciplines have a wealth of graph data~\cite{mason2007graph,harary1953graph,deo2017graph}, so that researchers can make full use of network analysis technologies to solve related problems in the corresponding fields. For example, in the field of life science, there are interactions between biological molecules, based on which the complex biological networks can be established and the molecular properties can thus be predicted~\cite{cunningham2020biophysical,madhukar2019bayesian}. In social science, the development of online social platforms has made it much easier to transmit information than ever before, but it is also followed by the wide spread of false news. Fortunately, by introducing graph data mining technologies, we can detect and filter the false information and purify the Internet environment to certain extent~\cite{vosoughi2018spread,nguyen2020fang}. In the field of network science, graph classification is one of the important graph data mining tasks. The representative methods of graph classification can be divided into three types, the first one is the graph kernel methods, which are mainly based on random walk~\cite{gartner2003graph}, path~\cite{hermansson2015generalized} and subtree~\cite{morris2017glocalized}; The second one is graph embedding methods, such as subgraph2vec~\cite{narayanan2016subgraph2vec} and graph2vec~\cite{narayanan2017graph2vec}; And the last one is deep learning methods, i.e., Graph Neural Networks (GNNs), which typically includes Dynamic Graph Convolutional Neural Networks (DGCNN)~\cite{zhang2018end}, DiffPool~\cite{diffpool} and SAGPool~\cite{lee2019self}.

For time series, here we mainly focus on modulation classification of radio signals~\cite{2018Big,8922798,8750803,zheng2019fusion}. In recent years, with the rapid development of deep learning, a number of neural network models have been introduced into this area~\cite{8446021,8330488,2020A}. Hong et al.~\cite{2017Automatic} proposed a classification model GRU based on two-layer Gated Recurrent Unit (GRU) network. Rajendran et al.~\cite{8357902} proposed a two-layer Long Short Term Memory (LSTM) model for automatic modulation classification, which has considerable accuracy but is relatively time-consuming. Some researchers mapped time series into matrices and then used Convolutional Neural Networks (CNNs) to classify them~\cite{8330488,8418751}. For instance, Wang et al.~\cite{wang2015encoding} transformed time series into matrices by adopting Gramian Angular Fields (GAF) and Markov Transition Fields (MTF). Peng et al.~\cite{8418751} obtained the constellation diagram of a signal, and then use the typical CNN for classification. However, since the mapping rules of these methods are fixed, they may not be flexible enough.
O’Shea et al.~\cite{o2016convolutional} treated each radio signal as a single-channel image with width 2 and used 2 convolution layers and 2 fully connected layers for modulation classification, we call this method CNN2D in this paper and use it as baseline for comparison. Besides, O’Shea et al.~\cite{o2018over} also used a residual neural network for modulation classification, namely CNN1D.

It is worth noting that, due to the great success of  GNN~\cite{2020Deep,9046288,DBLP:journals/corr/abs-1709-05584,8395024}, it would be also an exciting direction to first map time series into graphs and then utilize GNN to classify them.
Lacasa et al.~\cite{2008From} proposed the Visibility Graph (VG) algorithm, which can transform time series into graphs. Based on VG, Luque et al.~\cite{articlehvg} introduced Horizontal Visibility Graph (HVG), a mapping algorithm with simpler rules and fewer statistics in the transforming process. Gao et al.~\cite{2012Limited} proposed an improved algorithm, Limited Penetrable Visibility Graph (LPVG), which achieved good results in the classification of oil-gas-water three-phase flow. Xuan et al.~\cite{xuan2021clpvg} further introduced circle system for the first time in the construction of VG, proposed Circular Limited Penetrable Visibility Graph (CLPVG), and realized higher accuracy in radio modulation classification with graph embedding technology than LPVG. However, almost all of these techniques map time series into graphs according to fixed rules, and thus lack of flexibility to cope with complex tasks.

 In this paper, we focus on proposing an end-to-end GNN framework for signal classification. Particularly, we first introduce the Adaptive Visibility Graph (AVG), which can map signals into graphs adaptively through some neural network layers, thus avoiding the inflexibility of the mapping algorithms described above. Then we combine AVG with the modified DiffPool~\cite{diffpool}, a typical GNN model originally used for graph classification, to form an end-to-end classification framework, namely AVGNet. Different from the existing fixed mapping techniques, in our framework, the signal sample can be automatically and adaptively converted into the corresponding best matching graph, which not only greatly reduces the time consumed in the mapping process, but also can aggregate the local and global information of the signal sample. The main contributions of this paper are shown below.

\begin{itemize}
    \item We propose AVG, which can automatically and adaptively map signals into graphs according to the particular task. Through this mechanism, we can keep the local and global information of the original signals as much as possible, and can mine the implicit temporal information in the radio signals by analyzing graph topological features.
    \item We give an end-to-end GNN framework for radio signal modulation classification, namely AVGNet, by integrating AVG and the modified DiffPool. In this setting, AVG can be trained together with DiffPool, and adaptively learn to obtain the most suitable graph for the current signal, so as to significantly improve both the efficiency and effectiveness of the classification model.
    \item We testify our method on two public datasets in the field of radio signals, RML2016.10a and RML2016.10b, and the results validate the effectiveness of AVGNet, i.e., it outperforms RNN, CNN and traditional VG models, achieving the state-of-the-art performance.
\end{itemize}

The rest of paper is structured as follows. Sec.~\ref{sec_2} introduces AVG and how to combine it with DiffPool to build our end-to-end deep learning framework AVGNet. Sec.~\ref{sec_3} gives the datasets, simulation setting, baselines, and meanwhile summarizes and discusses the simulation results. Finally, the paper is concluded in Sec.~\ref{sec_4}.

\section{Method}
\label{sec_2}

\subsection{Mapping of Signals to Graphs\label{method_map}}
The classification problem of radio modulation modes can be expressed as an $M$-element hypothesis testing problem. Let the set of modulation classes be $\{1,2...,M\}$, where $M$  represent the category number. Then the modulation type of signal $T_{now}$ can be expressed as
\begin{equation}\label{eqadd_1}
\gamma(T_{now})=k,k\in[1,M],
\end{equation}
where $\gamma(T_{now})$ represents the modulation type of signal $T_{now}$.

\begin{figure}[!ht]
	\centering
    \includegraphics[scale=0.35]{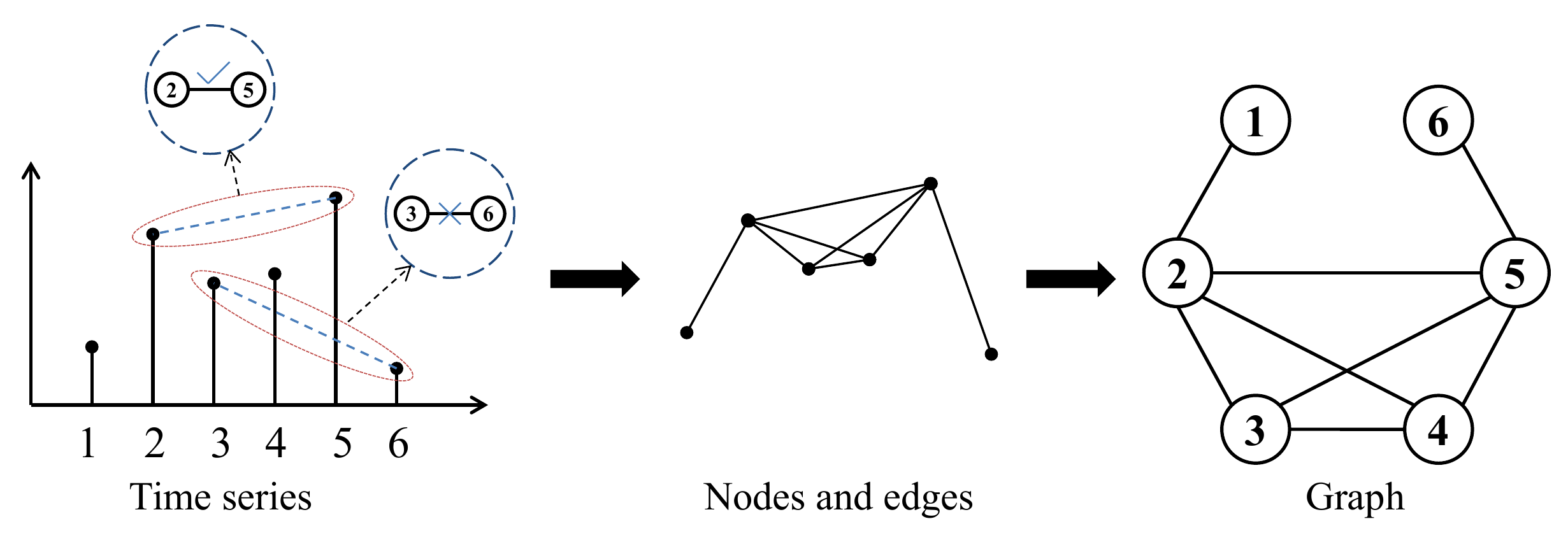}
	\caption{The process of mapping time series into a graph by VG model.}
	\label{fig:lpvg}
\end{figure}

In order to use the graph classification model in the field of GNN for radio signal modulation classification, we first need to convert the signals into graphs. Up to now, Lacasa et al.~\cite{2008From} proposed the typical Visibility Graph (VG) algorithm for mapping time series into graphs, whose process is shown in Fig.~\ref{fig:lpvg}. On this basis, Luque et al.~\cite{articlehvg} proposed the improved mapping algorithm, Horizontal Visibility Graph (HVG), and Gao et al.~\cite{2012Limited} further introduced Limited Penetrable Visibility Graph (LPVG). We take VG as an example to briefly introduce its rules for mapping signals into graphs. Given a signal $T$
\begin{equation}\label{eq1}
T=[T_1,T_2,\cdots,T_n],
\end{equation}
where $n$ represents the length of $T$, namely the number of sampling points, $T_n$ represents the signal value corresponding to sampling time point $n$. VG regards all sampling points as nodes of the final mapped graph, and then determines whether every two sampling points can form an edge. Taking sampling points $(i,T_i)$ and $(j,T_j)$ for example, these two nodes can form an edge if they satisfy the following condition: 
\begin{equation}\label{eqadd1}
T_u<\frac{(T_j-T_i)\times(u-j)}{(j-i)}+T_j, i<u<j,
\end{equation}
The specific process of VG mapping a signal into the corresponding graph is shown in Fig.~\ref{fig:lpvg}. It can be seen that the signal values corresponding to nodes 3 and 4 are all below the line segment formed by sampling points $(2,T_2)$ and $(5,T_5)$, so nodes 2 and 5 can form an edge. On the contrary, nodes 3 and 6 can not form an edge because the sampling point $(4,T_4)$ or $(5,T_5)$ between them is above their line segment.

\begin{figure*}[!t]
	\centering
    \includegraphics[scale=0.5]{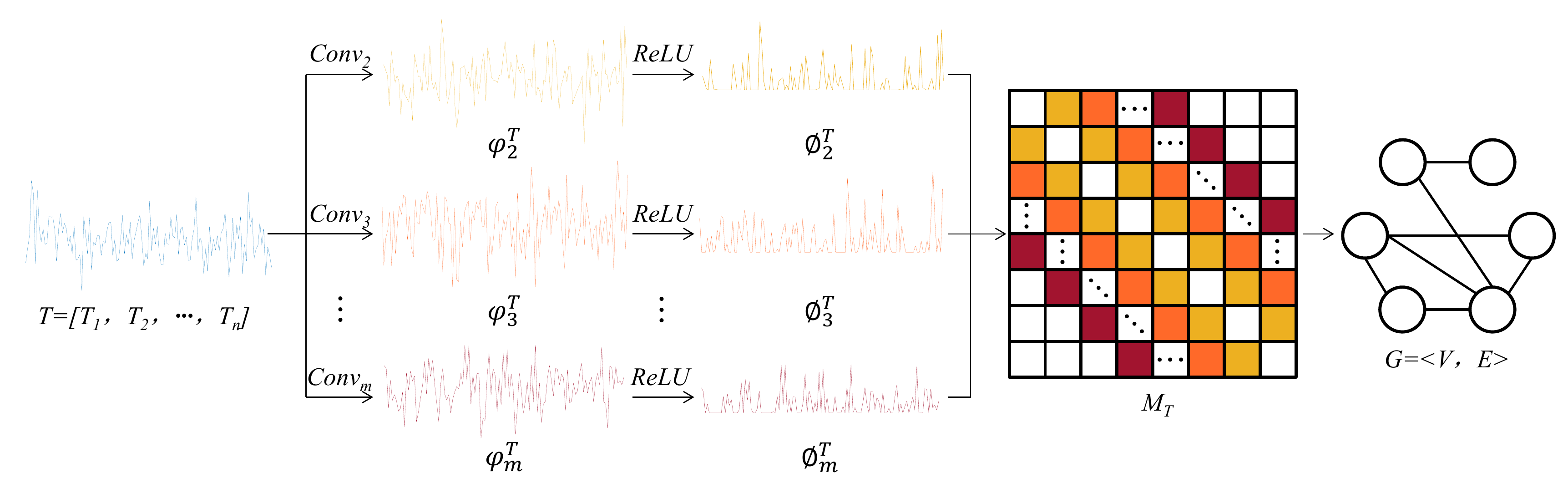}
	\caption{The process of mapping time series into a graph through AVG algorithm.}
	\label{fig:avg}
\end{figure*}

\begin{algorithm}[!t] 
  \caption{\textbf{The process of mapping time series to graph by Adaptive Visibility Graph (AVG)}}
  \label{alg1}
  \KwIn{Time series $T=[T_1,T_2,\cdots,T_n]$, hyperparameter $m$ shown in Eq.~\ref{eq2}.}
  \KwOut {Graph $G=\left<V, E\right>$ whose feature matrix is shown in Eq.~\ref{eq5}.}
  \For {s=2 to m}
  { 
    Define one-dimensional convolution layer $Conv_s(\cdot)$ whose convolution kernel length is $s$ and step length is 1 \\
    Compute the feature sequence $\phi_s^T=ReLU(Conv_s(T))=[F_1^s,F_2^s,\cdots,F_f^s,\cdots,F_{n+1-s}^s],s\in[2,m]$ \\
  }
  Obtain the feature matrix $M_T$ shown in Eq.~\ref{eq5}\\
  return mapped graph $G=\left<V, E\right>$
\end{algorithm}

In general, the only difference between the above mapping methods is their fixed mapping rules. Although all of these mapping algorithms can map time series samples into graphs so that the knowledge of graph domain can be reasonably introduced to solve downstream tasks in time series problems, it is undeniable that these mapping algorithms all need to judge whether each two sampling points can form an edge of the graph successively, which is relatively inefficient, and the required mapping time increases multiply with the length of the time series. In addition, these mapping methods are relatively rigid and can only build graphs based on established rules, which lacks of flexibility.

In order to solve the above problems, and retain the information of original signals as much as possible in the mapping process and further improve the mapping efficiency, we propose the Adaptive Visibility Graph (AVG) algorithm on the basis of the above existing mapping methods~\cite{2008From},~\cite{articlehvg},~\cite{2012Limited}. AVG can transform time series into graphs through one-dimensional convolution, and then can be trained together with the typical graph classification models. In general, the specific process of this algorithm is shown in Fig.~\ref{fig:avg} and Algorithm~\ref{alg1}. For the time series $T$ shown in Eq.~(\ref{eq1}), we aim to map it into a graph $G=\left<V, E\right>$, where $V$ and $E$ represent node set and edge set of graph $G$, respectively. At first, we use $m-1$ one-dimensional convolution operators $Conv_s$ with different kernel length $s\in[2, m]$ to process $T$. The obtained feature sequences of $T$ can be calculated by
\begin{equation}\label{eq2}
\varphi_s^T=Conv_s (T),s\in[2,m],
\end{equation}
with $\varphi_s^T$ being vectors expressed as
\begin{equation}\label{eq3}
\varphi_s^T=[B_1^s,B_2^s,\cdots,B_{n+1-s}^s],s\in[2,m].
\end{equation}
Here, the step length of convolution operator is set to 1, and $m$ is a hyperparameter used to limit the distance between two time sampling points. After that, we use the nonlinear activation function ReLU~\cite{2011Deep} to process $\varphi_s^T$, where ReLU can be similar to the operation of filtering noise there. The process can be represented as
\begin{equation}
\begin{aligned}
\label{eq4}
	\phi_s^T&=ReLU(\varphi_s^T)\\
	&=[F_1^s,F_2^s,\cdots,F_f^s,\cdots,F_{n+1-s}^s],s\in[2,m].
\end{aligned}
\end{equation}
As a result, the univariate time series $T$ shown in Eq.~(\ref{eq1}) can be mapped into the graph $G=\left<V, E\right>$ which can be expressed by a feature matrix
\begin{small}
\begin{equation}\label{eq5}
M_T\!=\!
{\left[ \begin{array}{cccccccc}
\!0\!&\!F_1^2\!&\!F_1^3\!&\!\cdots\!&\!F_1^m\!&\!0\!&\!\cdots\!&\!0\! \\
\!F_1^2\!&\!0\!&\!F_2^2\!&\!F_2^3\!&\!\cdots\!&\!F_2^m\!&\!\ddots\!&\!\vdots\! \\
\!F_1^3\!&\!F_2^2\!&\!0\!&\!F_3^2\!&\!F_3^3\!&\!\ddots\!&\!\ddots\!&\!0\! \\
\!\vdots\!&\!F_2^3\!&\!F_3^2\!&\!0\!&\!F_4^2\!&\!F_4^3\!&\!\ddots\!&\!F_{n+1-m}^m\! \\
\!F_1^m\!&\!\vdots\!&\!F_3^3\!&\!F_4^2\!&\!0\!&\!F_5^2\!&\!\ddots\!&\!\vdots\! \\
\!0\!&\!F_2^m\!&\!\ddots\!&\!F_4^3\!&\!F_5^2\!&\!0\!&\!\ddots\!&\!F_{n-2}^3\! \\
\!\vdots\!&\!\ddots\!&\!\ddots\!&\!\ddots\!&\!\ddots\!&\!\ddots\!&\!\ddots\!&\!F_{n-1}^2\! \\
\!0\!&\!\cdots\!&\!0\!&\!F_{n+1-m}^m\!&\!\cdots\!&\!F_{n-2}^3\!&\!F_{n-1}^2\!&\!0\!
\end{array} 
\right ]}
\end{equation}
\end{small}

Generally, AVG algorithm is inspired by VG and put forward on the basis of it. When determining whether two signal sampling points can construct an edge, VG algorithm will first establish a linear function of the form $y=kx+b$ based on the signal values of these two sampling points, and then check whether the signal values in between are all below the line, as shown in Fig.~\ref{fig:lpvg}. Such process may ignore the specific signal values in between, which will inevitably lead to the loss of some information. In contrast, when judging whether two signal points can form an edge of graph, our proposed AVG algorithm can make full use of the signal values corresponding to the two and the time points in between through the one-dimensional convolution layers. In particular, the one-dimensional convolution operation in AVG Algorithm is similar to the linear function involved in the VG algorithm, which is used to measure the correlation between the sliding signal segments and the convolution operator as an important temporal feature. Note that the maximum length of the signal segments can be limited by the hyperparameter $m$ in Eq.~(\ref{eq2}), indicating that two signal sampling points whose time interval is greater than $m$ can not form an edge, because in the time series, the temporal information of two time sampling points that are far apart is not as important as those close to each other. By focusing on the local relevance of the signal, we can also speed up the mapping efficiency. As a result, AVG algorithm can efficiently keep as much information in the time series as possible, which is beneficial to the subsequent classification task.

\subsection{Classification Combined with GNN}
After mapping radio signals into graphs by means of the proposed AVG algorithm, we can combine the typical GNN model used for graph classification to obtain a neural network framework for radio signal modulation classification. This classification framework can greatly improve the mapping efficiency with the help of one-dimensional convolution and has excellent classification effect. In this paper, we choose DiffPool from the existing graph classification models in GNN domain. It is worth noting that DiffPool requires the feature vector for each node in addition to the adjacency matrix or feature matrix of the input graph. On the one hand, for common univariate time series as shown in~(\ref{eq1}), we can get the feature matrix of mapped graph through AVG, and take the signal value of the sampling point corresponding to the node as node's feature sequence, so that we are able to directly combine AVG and DiffPool to form a classification model to realize time series classification. On the other hand, as for the radio signal studied in this paper, one of the typical multivariable time series, we need to process the DiffPool to classify the input multi-channel samples. Each signal sample is composed of channels I and Q, and let the two channels be represented as
\begin{equation}\label{eq6}
I=[I_1,I_2,\cdots,I_j,\cdots,I_n],
\end{equation}
\begin{equation}\label{eq7}
Q=[Q_1,Q_2,\cdots,Q_j,\cdots,Q_n].
\end{equation}

We use AVG algorithm to map the channel I and channel Q of each signal sample into their corresponding graphs $G_I=\left<V_I,E_I\right>$ and $G_Q=\left<V_Q,E_Q\right>$, respectively, that is, get the feature matrices $M_I$ and $M_Q$ of mapped graph through
\begin{equation}
\begin{aligned}
\label{eq8}
\phi_s^I&=ReLU(Conv_s(I)) \\
&=[J_1^s,J_2^s,\cdots,J_{n+1-s}^s],s\in[2,m],
\end{aligned}
\end{equation}

\begin{equation}
\begin{aligned}
\label{eq9}
\phi_s^Q&=ReLU(Conv_s(Q)) \\
&=[P_1^s,P_2^s,\cdots,P_{n+1-s}^s],s\in[2,m],
\end{aligned}
\end{equation}
where $M_I$ and $M_Q$ can be represented as
\begin{small}
\begin{equation}\label{eq10}
M_I\!=\!
{\left[ \begin{array}{cccccccc}
\!0\!&\!J_1^2\!&\!J_1^3\!&\!\cdots\!&\!J_1^m\!&\!0\!&\!\cdots\!&\!\!\!\!0\! \\
\!J_1^2\!&\!0\!&\!J_2^2\!&\!J_2^3\!&\!\cdots\!&\!J_2^m\!&\!\ddots\!&\!\!\!\!\vdots\! \\
\!J_1^3\!&\!J_2^2\!&\!0\!&\!J_3^2\!&\!J_3^3\!&\!\ddots\!&\!\ddots\!&\!\!\!\!0\! \\
\!\vdots\!&\!J_2^3\!&\!J_3^2\!&\!0\!&\!J_4^2\!&\!J_4^3\!&\!\ddots\!&\!\!\!\!J_{n+1-m}^m\! \\
\!J_1^m\!&\!\vdots\!&\!J_3^3\!&\!J_4^2\!&\!0\!&\!J_5^2\!&\!\ddots\!&\!\!\!\!\vdots\! \\
\!0\!&\!J_2^m\!&\!\ddots\!&\!J_4^3\!&\!J_5^2\!&\!0\!&\!\ddots\!&\!\!\!\!J_{n-2}^3\! \\
\!\vdots\!&\!\ddots\!&\!\ddots\!&\!\ddots\!&\!\ddots\!&\!\ddots\!&\!\ddots\!&\!\!\!\!J_{n-1}^2\! \\
\!0\!&\!\cdots\!&\!0\!&\!J_{n+1-m}^m\!&\!\cdots\!&\!J_{n-2}^3\!&\!J_{n-1}^2\!&\!\!\!\!0\!
\end{array} 
\right ]}
\end{equation}
\end{small}

\begin{small}
\begin{equation}\label{eq11}
M_Q\!=\!
{\left[ \begin{array}{cccccccc}
\!0\!&\!P_1^2\!&\!P_1^3\!&\!\cdots\!&\!P_1^m\!&\!0\!&\!\cdots\!&\!\!\!\!0\! \\
\!P_1^2\!&\!0\!&\!P_2^2\!&\!P_2^3\!&\!\cdots\!&\!P_2^m\!&\!\ddots\!&\!\!\!\!\vdots\! \\
\!P_1^3\!&\!P_2^2\!&\!0\!&\!P_3^2\!&\!P_3^3\!&\!\ddots\!&\!\ddots\!&\!\!\!\!0\! \\
\!\vdots\!&\!P_2^3\!&\!P_3^2\!&\!0\!&\!P_4^2\!&\!P_4^3\!&\!\ddots\!&\!\!\!\!P_{n+1-m}^m\! \\
\!P_1^m\!&\!\vdots\!&\!P_3^3\!&\!P_4^2\!&\!0\!&\!P_5^2\!&\!\ddots\!&\!\!\!\!\vdots\! \\
\!0\!&\!P_2^m\!&\!\ddots\!&\!P_4^3\!&\!P_5^2\!&\!0\!&\!\ddots\!&\!\!\!\!P_{n-2}^3\! \\
\!\vdots\!&\!\ddots\!&\!\ddots\!&\!\ddots\!&\!\ddots\!&\!\ddots\!&\!\ddots\!&\!\!\!\!P_{n-1}^2\! \\
\!0\!&\!\cdots\!&\!0\!&\!P_{n+1-m}^m\!&\!\cdots\!&\!P_{n-2}^3\!&\!P_{n-1}^2\!&\!\!\!\!0\!
\end{array} 
\right ]}
\end{equation}
\end{small}

\begin{figure*}[!t]
	\centering
    \includegraphics[scale=0.5]{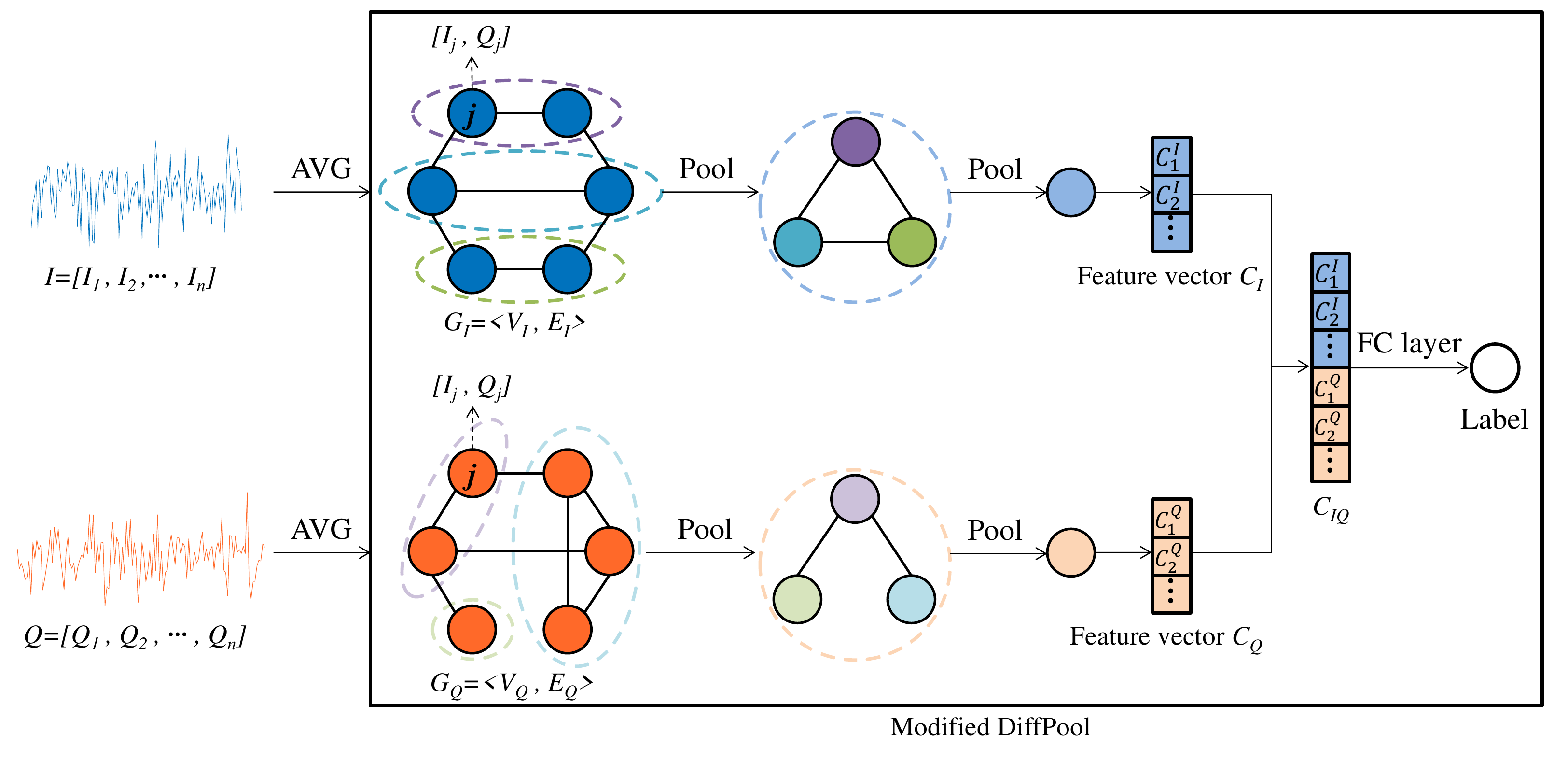}
	\caption{The structure of the proposed classification framework AVGNet.}
	\label{fig_AVGNet}
\end{figure*}

Then, we set the feature vectors of the respective nodes $j$ of graphs $G_I$ and $G_Q$ as $[I_j,Q_j]$, and then use the modified DiffPool without the last fully connected layer to process graphs $G_I$ and $G_Q$, respectively, to get their respective feature vectors $C_I$ and $C_Q$:
\begin{equation}\label{eq12}
C_I=[C_1^I,C_2^I,\cdots],
\end{equation}
\begin{equation}\label{eq13}
C_Q=[C_1^Q,C_2^Q,\cdots].
\end{equation}
After combining the feature sequences $C_I$ and $C_Q$ into $C_{IQ}$ through
\begin{equation}
\begin{aligned}
\label{eq14}
C_{IQ}&=concatenate\left<C_I, C_Q\right> \\
&=[C_1^I,C_2^I,\cdots,C_1^Q,C_2^Q,\cdots],
\end{aligned}
\end{equation}
we add a new fully connected layer for classification. Finally, we can get the classification model called Adaptive Visibility Graph GNN Framework (AVGNet), which is used to classify the radio signals.

It is obvious that compared with the existing traditional visible mapping algorithms, our AVG mapping algorithm can be trained with the typical GNN to form an end-to-end modulation signal classification model. The model can learn to map radio signal into a more appropriate graph by neural network training. By mapping signals into graphs and then classifying them, the temporal information of signal data can be fully preserved through the topological structure of graph. In addition, we can express the degree of association between time sampling points corresponding to the graph nodes through the feature matrix of graph, namely the weight matrix of edges such as $M_I$. At the same time, we are able to retain the signal values of the time sampling points in the original signal data with the help of the feature vectors of mapped graph nodes.

\subsection{The Overall Framework}
The overall framework of AVGNet is shown in Fig.~\ref{fig_AVGNet}. Through AVG algorithm, we can map the signal data of I and Q channels into graphs $G_I$ and $G_Q$, respectively. After setting the feature vectors of nodes of graphs $G_I$ and $G_Q$, we use the modified DiffPool, a typical GNN model for graph classification, to process graphs $G_I$ and $G_Q$, respectively, and get their corresponding feature vectors $C_I$ and $C_Q$. After concatenating $C_I$ and $C_Q$ and then getting $C_{IQ}$, we can achieve the goal of classification by processing $C_{IQ}$ through a fully connected layer.

Generally, by combining AVG algorithm with the modified DiffPool, we can obtain the AVGNet model for radio signal modulation classification. In particular, the part of mapping signals into graphs and graph classification are trained together, which is able to ensure that the obtained graph is more appropriate for the original signal. Therefore, our AVGNet model can retain the information in the original signal as much as possible and dig out the potential information to improve the classification accuracy.

\section{SIMULATIONS}
\label{sec_3}

\subsection{Datasets}
In this paper, we use two public radio signal datasets RML2016.10a~\cite{o2016convolutional} and RML2016.10b~\cite{grcon} to testify our classification model AVGNet.

Among them, RML2016.10a has a total of 220,000 signal samples, including 8PSK, AM-DSB, AM-SSB, BPSK, CPFSK, GFSK, PAW4, QAM16, QAM64, QPSK and WBFM in 11 modulation types. Each modulation type has 20 levels of signal-to-noise ratios (SNRs) ranging from -20dB to 18dB at 2dB intervals. Each signal sample is 128 in length and consists of two channels I and Q. In the simulations, we divide the training set and the test set in a ratio of 4:1. In order to ensure the balance of the number of samples, we randomly select 80\% signal data from each SNR of each modulation mode as the training set, and the rest as the test set. Finally, there are 176,000 signal samples in the training set and 44,000 signal samples in the test set.

In addition, the dataset RML2016.10b has 1.2 million samples and contains 10 modulation types, namely 8PSK, AM-DSB, BPSK, CPFSK, GFSK, PAM4, QAM16, QAM64, QPSK and WBFM. Again, the dataset has 20 levels of SNRs and a length of 128. In the simulations, we divide this dataset according to the open source method in ~\cite{o2016convolutional}, that is, we take the random seed value as 2016, and then randomly divide it into training set and test set in a 1:1 ratio.

\subsection{Simulation Setting}
In the simulations, we use cross-entropy loss function and Adam~\cite{kingma2014adam} optimizer. The batch size and initial value of learning rate are set to 128 and 0.001, respectively, and the learning rate decreases to 80\% of the previous value for every 10 training batches. In addition, the hyperparameter $m$ mentioned in Eq.~\ref{eq3} of our proposed model AVGNet is set to 11. All simulations are run on NVIDIA Tesla V100 based on the PyTorch deep learning framework~\cite{paszke2017automatic}. What's more, we use PyTorch Geometric~\cite{Fey/Lenssen/2019}, a geometric deep learning extension library for PyTorch, to build the neural network layers of DiffPool in our AVGNet model.

\subsection{Baselines}
We compare AVGNet with three different types of baselines. The first is to use the RNN model to classify the signals, specifically, we use GRU and LSTM mentioned in Sec.~\ref{introduction}. The second is to classify radio signals with the help of CNN models. In the simulation, we mainly use GAF, MTF, CD, CNN2D and CNN1D mentioned in Sec.~\ref{introduction}. In particular, the first three methods GAF, MTF and CD map radio signals into images and then classify them using the typical ResNet50 model. The last one is to convert signals into graphs by VG mapping method with fixed conversion rules, and then classify the signals by classifying the corresponding graphs. In the mapping methods, we choose the typical methods VG described in Sec.~\ref{method_map} and LPVG. In order to ensure the fairness of comparison, we also use the modified DiffPool in AVGNet to classify the obtained graphs.

\subsection{Results and Discussion}
\begin{figure}[!t]
	\centering
    \includegraphics[scale=0.5]{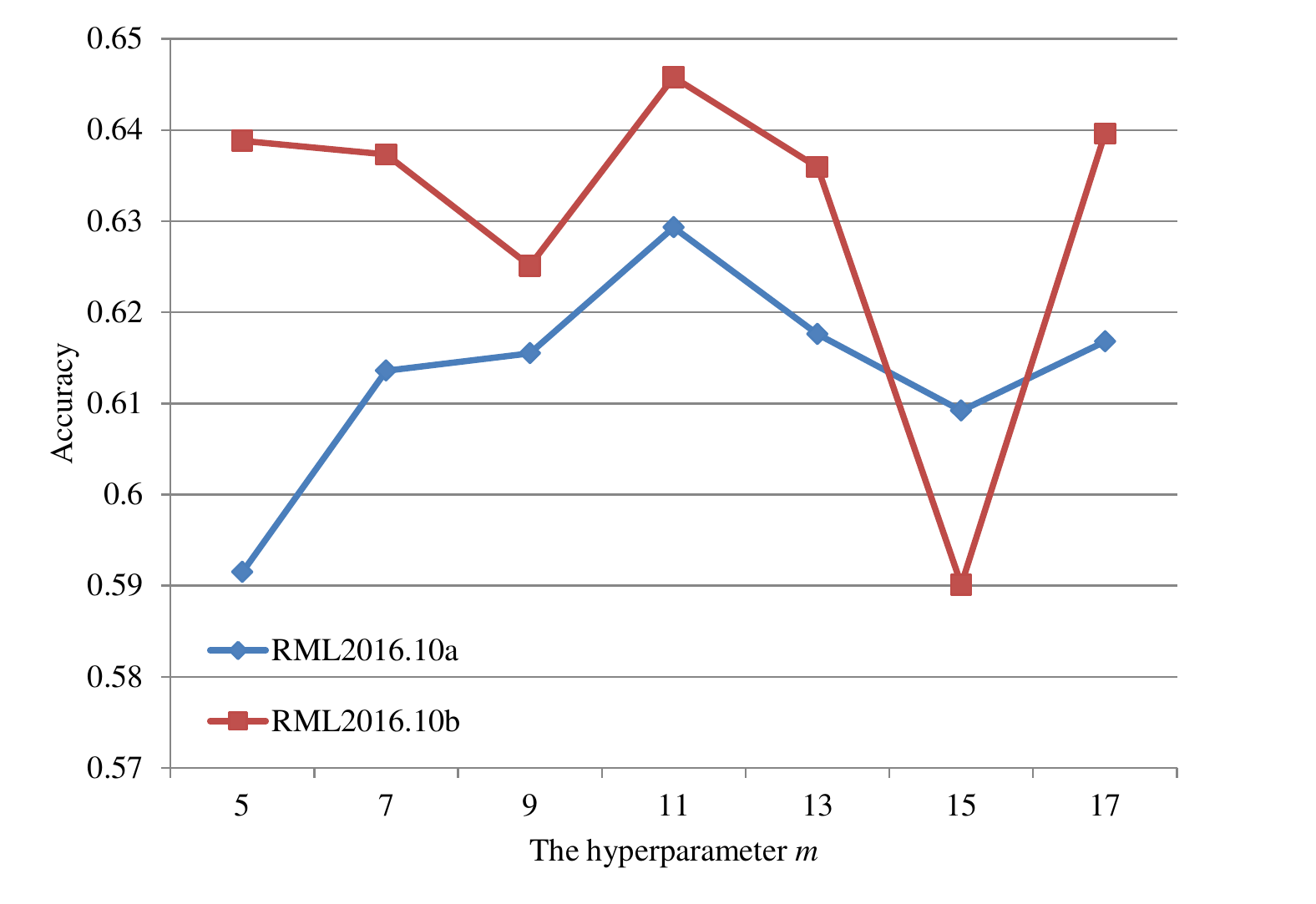}
	\caption{The classification accuracy of different hyperparameter $m$.}
	\label{fig_m}
\end{figure}

At first, we briefly discuss the effect of the hyperparameter $m$ mentioned in~(\ref{eq3}) on classification accuracy. In the case of different hyperparameter $m$, the classification accuracy of AVGNet on the two datasets is shown in Fig.~\ref{fig_m}. It can be seen that when $m$ is equal to 11, the classification accuracy of AVGNet on the two datasets is the best.

\begin{figure*}[!t]
    \centering
    \subfigure[RML2016.10a]{
        \includegraphics[width=0.48\textwidth]{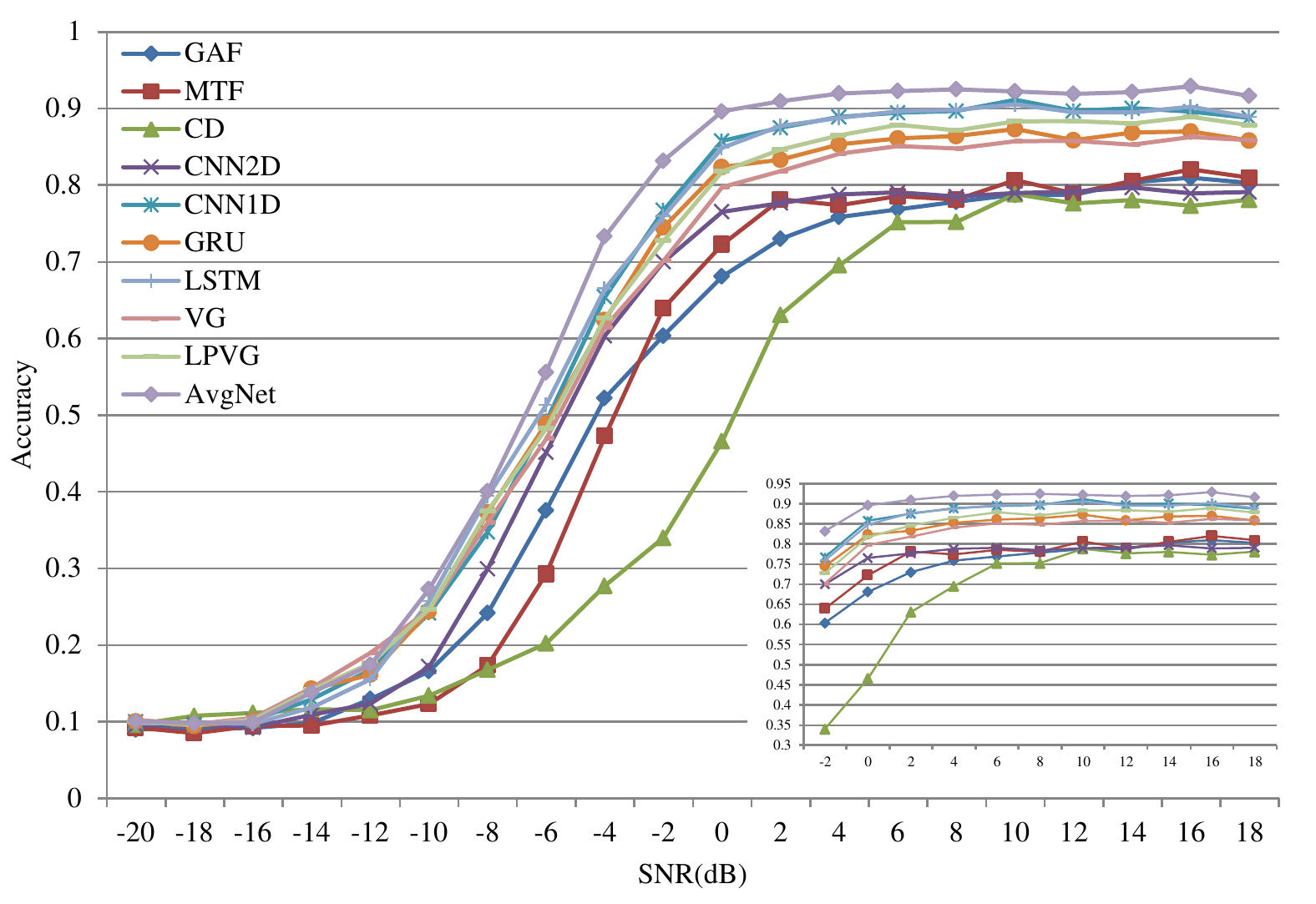}
    }
    \subfigure[RML2016.10b]{
        \includegraphics[width=0.48\textwidth]{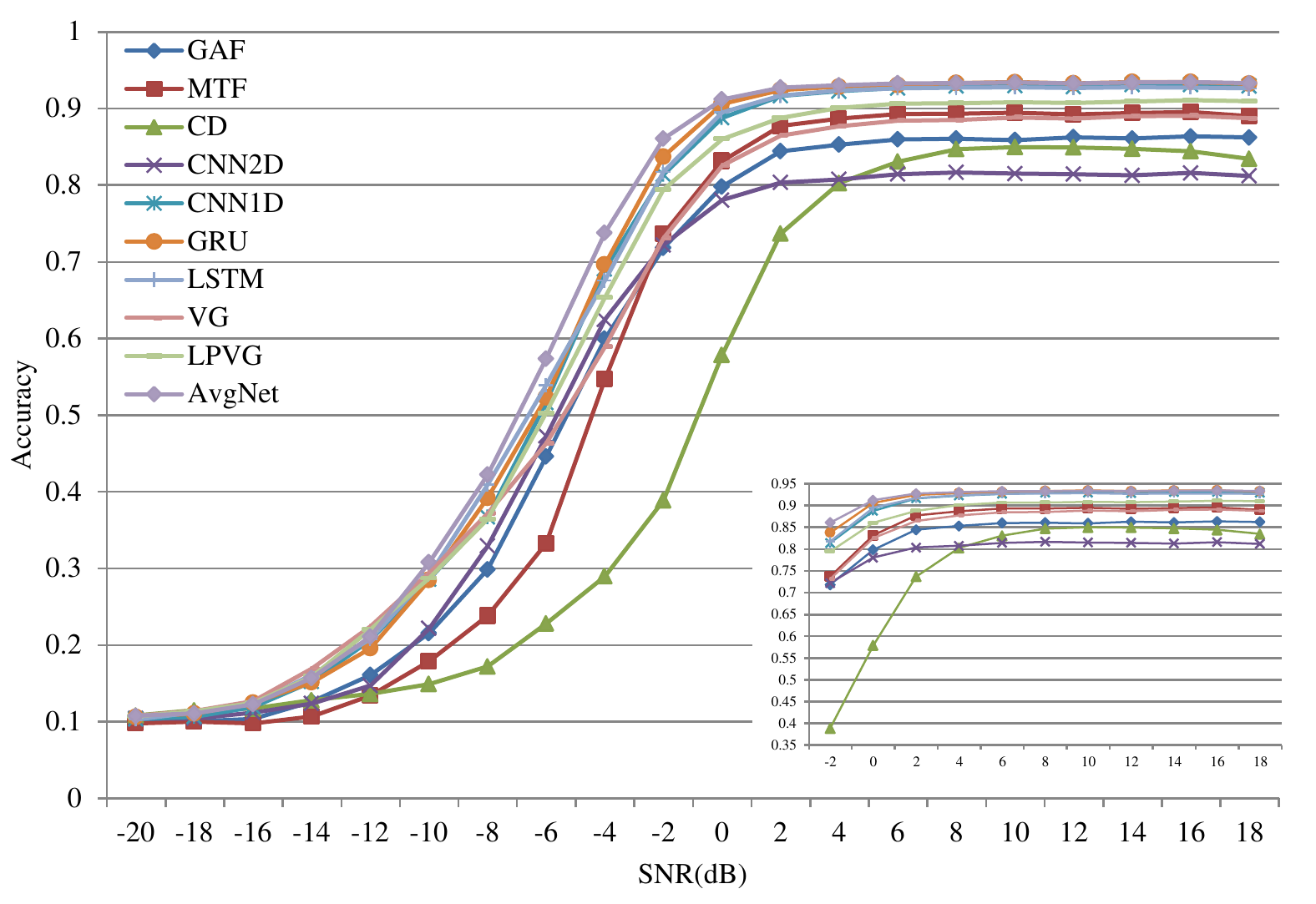}
    }
    \caption{The classification accuracy of different methods and (a) RML2016.10a, (b) RML2016.10b.}
    \label{fig_snr}
\end{figure*}

Then, we mainly compare the classification effect of AVGNet and different types of baselines on radio signal datasets. The classification results such as accuracy, F1 score and recall rate obtained by all the methods we used for datasets RML2016.10a and RML2016.10b are shown in TABLE~\ref{table10a} and TABLE~\ref{table10b}, respectively, and their classification accuracy on different SNRs of the corresponding test set are shown in Fig.~\ref{fig_snr} (a) and Fig.~\ref{fig_snr} (b), respectively. We first compare AVGNet with the typical RNN models GRU and LSTM used for radio signal modulation classification. As can be seen from TABLE~\ref{table10a} and Fig.~\ref{fig_snr} (a), AVGNet performs better than both GRU and LSTM on dataset RML2016.10a in accuracy, F1 score and recall rate. And the classification accuracy of AVGNet is higher than that of GRU and LSTM in each SNR. In addition, AVGNet also outperforms the CNN methods such as GAF, MTF, CD, CNN2D and CNN1D. It is worth noting that the DiffPool used in AVGNet and the ResNet50 used in GAF, MTF and CD are both typical deep learning models for classification in the graph and image domains, respectively. They both have good classification performance in their respective domains. To certain extent, it can be explained that, compared with mapping signals to images, mapping signals to graphs could help us to better capture the latent features of signals, leading to higher signal classification accuracy. Finally, by comparing AVGNet with the typical graph mapping methods VG and LPVG, both of which use the modified DiffPool to classify the mapped graphs, we can find that AVG is better than VG and LPVG with fixed transformation rules. In other words, AVG can learn more suitable graphs of signals through end-to-end training, leading to better classification performance. In general, our proposed AVGNet has the best classification performance among all the compared methods, with the reasonable model size. Moreover, for the other dataset RML2016.10b, we can also draw the same conclusion through TABLE~\ref{table10b} and Fig.~\ref{fig_snr} (b), indicating that our AVGNet has strong generalization ability.

\begin{table}[!t]
\centering
\caption{The classification performance of different methods on datasetset RML2016.10a.}
\begin{tabular}{c|cccc}
\hline
Method & Accuracy & F1 Score & Recall & Params size (MB)  \\ \hline 
GAF & 50.58\% & 0.5256 & 0.5058 & 94.11 \\ 
MTF & 50.26\% & 0.5342 & 0.5026 & 94.11 \\ 
CD & 44.32\% & 0.4639 & 0.4432 & 94.10 \\ 
CNN2D & 53.03\% & 0.5371 & 0.5303 & 10.83 \\
CNN1D & 60.02\% & 0.6193 & 0.6002 & 0.40 \\
GRU & 58.20\% & 0.5910 & 0.5820 & 2.45 \\
LSTM & 60.26\% & 0.6197 & 0.6026 & 0.82 \\ 
VG & 57.37\% & 0.5865 & 0.5751 & 1.82 \\
LPVG & 58.85\% & 0.6032 & 0.5899 & 1.82 \\
\textbf{AVGNet} & \textbf{62.93\%} & \textbf{0.6522} & \textbf{0.6293} & 1.83 \\ \hline
\end{tabular}
\label{table10a}
\end{table}

\begin{table}[!t]
\centering
\caption{The classification performance of different methods on datasetset RML2016.10b.}
\begin{tabular}{c|cccc}
\hline
Method & Accuracy & F1 Score & Recall & Params size (MB)  \\ \hline 
GAF & 56.96\% & 0.5924 & 0.5699 & 94.10 \\ 
MTF & 57.10\% & 0.6064 & 0.5713 & 94.10 \\ 
CD & 49.26\% & 0.5108 & 0.4931 & 94.09 \\ 
CNN2D & 55.27\% & 0.5566 & 0.5531 & 10.83 \\
CNN1D & 62.97\% & 0.6274 & 0.6301 & 0.40 \\
GRU & 63.57\% & 0.6358 & 0.6361 & 2.45 \\
LSTM & 63.33\% & 0.6337 & 0.6336 & 0.79 \\ 
VG & 59.86\% & 0.6029 & 0.5989 & 1.82 \\
LPVG & 61.67\% & 0.6146 & 0.6170 & 1.82 \\
\textbf{AVGNet} & \textbf{64.58\%} & \textbf{0.6469} & \textbf{0.6461} & 1.83 \\ \hline
\end{tabular}
\label{table10b}
\end{table}

\begin{figure*}[!t]
    \centering
    \subfigure[RML2016.10a]{
        \includegraphics[width=0.98\textwidth]{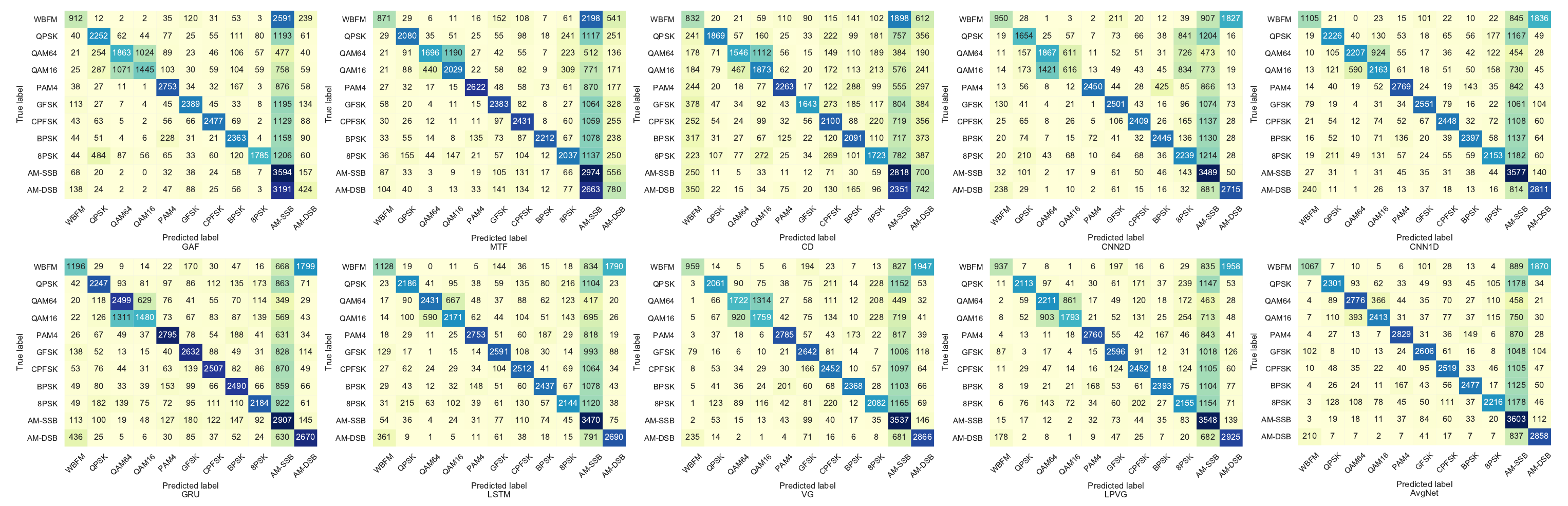}
    }
    \subfigure[RML2016.10b]{
        \includegraphics[width=0.98\textwidth]{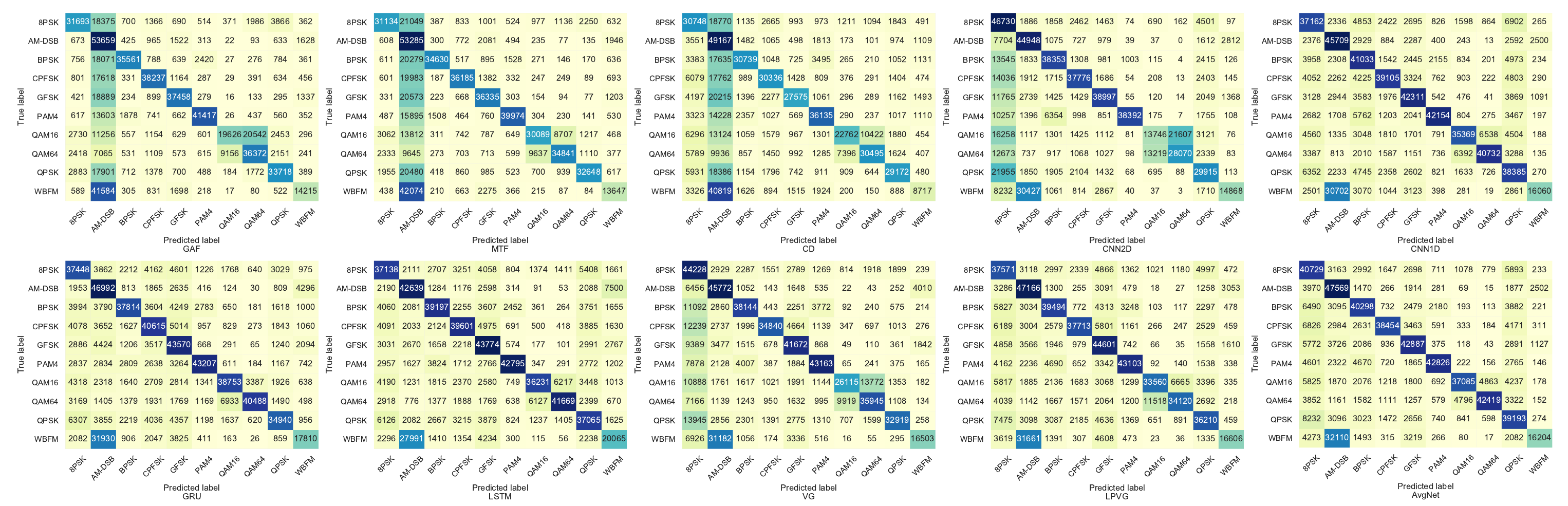}
    }
    \caption{The confusion matrices of different methods and (a) RML2016.10a, (b) RML2016.10b.}
    \label{fig_confusion}
\end{figure*}

\begin{figure*}[!t]
    \centering
    \subfigure[RML2016.10a]{
        \includegraphics[width=0.98\textwidth]{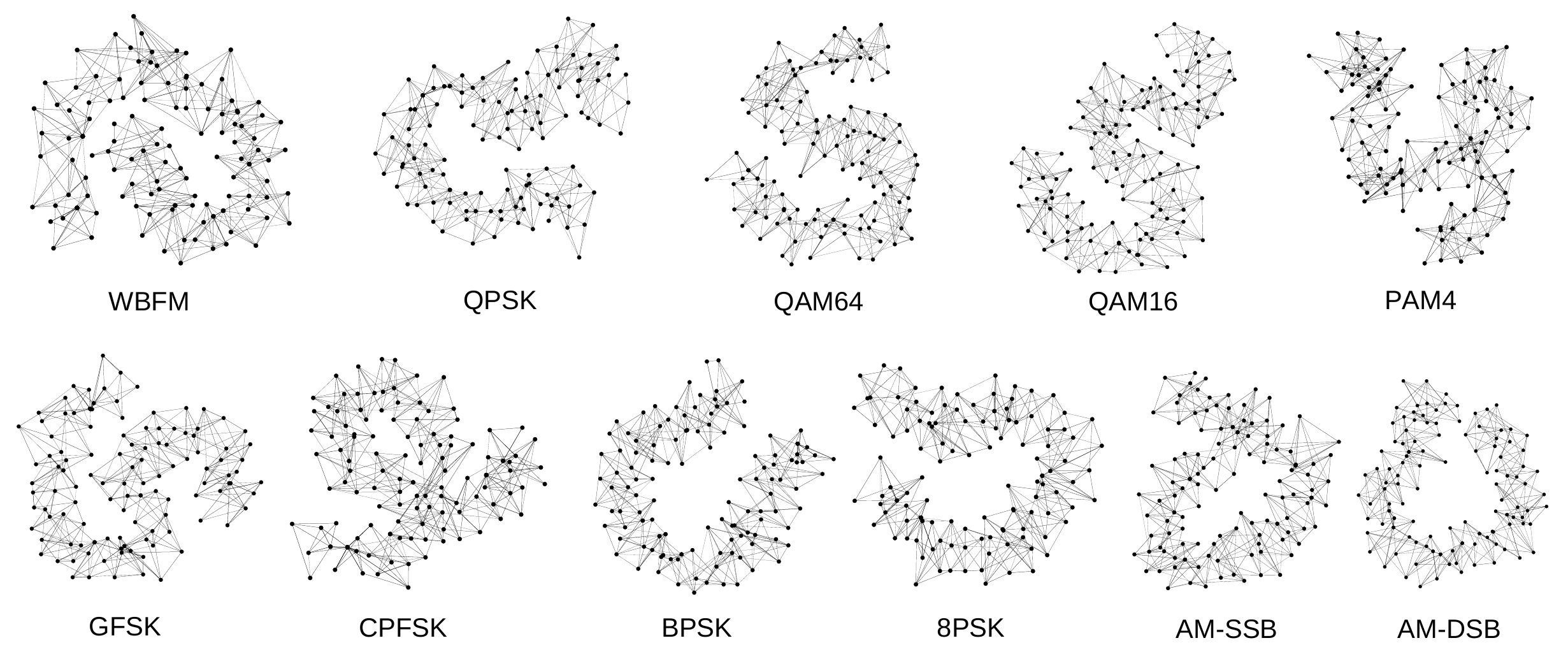}
    }
    \subfigure[RML2016.10b]{
        \includegraphics[width=0.98\textwidth]{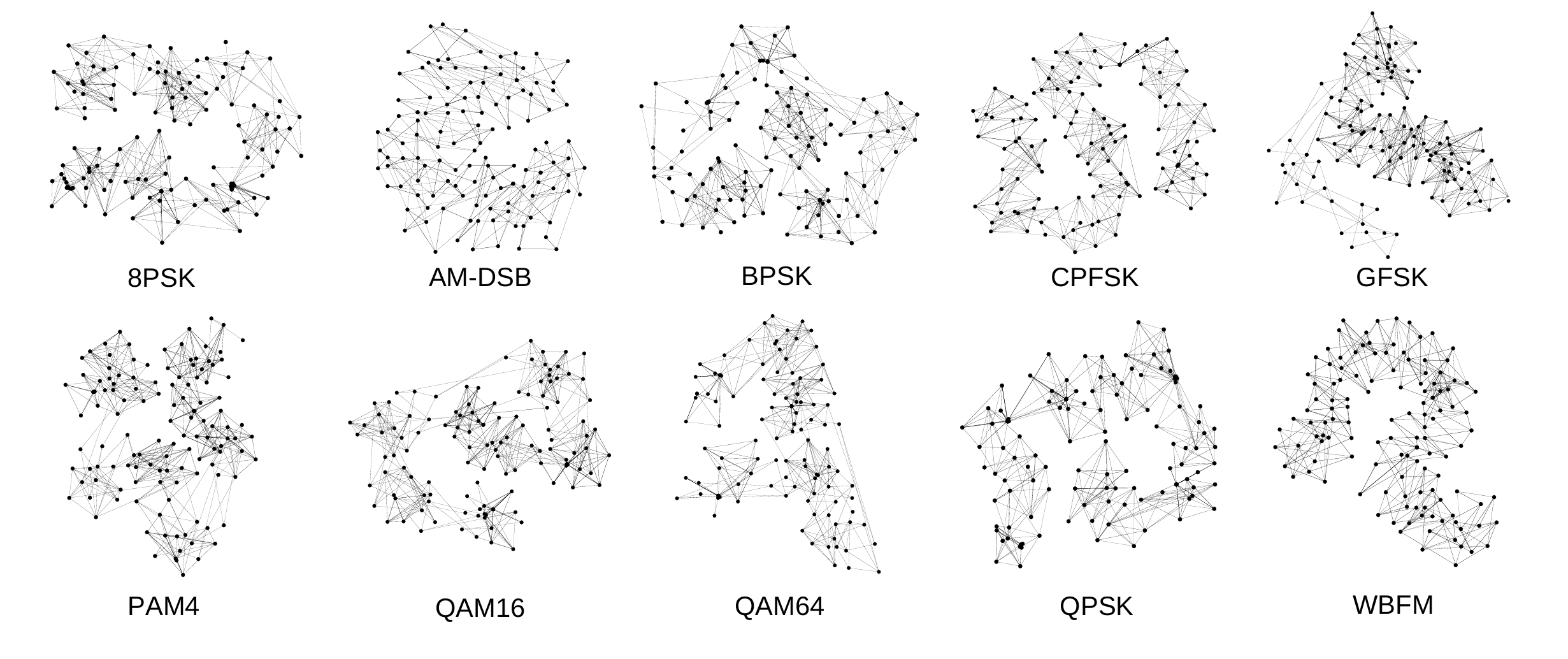}
    }
    \caption{The mapped graphs in AVGNet for (a) RML2016.10a, and (b) RML2016.10b.}
    \label{fig_graph}
\end{figure*}

In order to compare the classification performance of these different methods in more details, we plot their confusion matrices on the datasets RML2016.10a and RML2016.10b, as shown in Fig.~\ref{fig_confusion} (a) and Fig.~\ref{fig_confusion} (b), respectively. On the dataset RML2016.10a, all the methods tend to misclassify other modulation types as AM-SSB. Except for AVGNet, other classification methods are more likely to confuse the modulation types QAM16 and QAM64. Except GAF, MTF and CD, other methods are easy to misclassify WBFM as AM-DSB. On the dataset RML2016.10b, all methods are easy to misidentify WBFM as AM-DSB, in addition, GAF, MTF and CD, which map signals to images, are more likely to misclassify other modulation types as AM-DSB. Compared with other methods, CNN1D, GRU, LSTM and AVGNet have better ability to distinguish QAM16 and QAM64.

Furthermore, it is also very interesting to investigate the graphs generated by AVG in AVGNet. We randomly select a high SNR I-channel signal from each modulation type of the two datasets and draw their corresponding graphs, as shown in Fig.~\ref{fig_graph} (a) and Fig.~\ref{fig_graph} (b), respectively. These graphs are drawn by Gephi~\cite{Bastian09gephi:an}, and the layout is OpenOrd~\cite{Bastian09gephi:an}. According to Fig.~\ref{fig_graph} (a), on the dataset RML2016.10a, we find that most of the graphs mapped from different modulation types of signals present different structure, which is conducive to identifying the modulation types of radio signals. In addition, we can also see that although some graphs corresponding to different modulation types are similar in structure, such as QAM64 and GFSK, BPSK and AM-DSB, our proposed AVGNet can still distinguish them well, since the input of the GNN model DiffPool used in AVGNet not only contains the adjacency matrix representing the graph structure, but also contains the feature vector of each node in the graph. Similarly, according to Fig.~\ref{fig_graph} (b), we can find the similar situation on the dataset RML2016.10b. The graphs mapped from the modulation types QAM64 and QPSK are similar in structure, but we can still identify them pretty well. In general, we set the node feature vectors of the mapped graph in AVGNet based on the signal values of I and Q channels, and then make full use of them through the structure of the generated graph. In this sense, our AVGNet can comprehensively consider the local and global information of the original signal as a whole, leading to its outstanding classification performance.

\section{CONCLUSION}
\label{sec_4}
In this paper, we propose a new deep learning framework based on graph neural network, namely AVGNet, for radio signal modulation classification, which first maps the radio signals into graphs by using the trainable AVG algorithm, and then classifies them by the modified DiffPool. In particular, the process of mapping signals into graphs is trained together the graph classifier. And the simulation results show that our AVGNet has surprisingly good classification performance. In the future work, we may try to combine other graph classification models with AVG, or combine the graph data enhancement methods such as subgraph networks~\cite{8924759} in the field of network science, to further improve the classification performance of AVGNet. Meanwhile, we will also adopt AVGNet for the classification of other time series beyond radio signals, to validate its generalization ability across domains.

\ifCLASSOPTIONcaptionsoff
  \newpage
\fi



\bibliographystyle{IEEEtran}

\end{document}